\documentclass[10pt,twocolumn,letterpaper]{article}

\usepackage[pagenumbers]{cvpr} 

\usepackage{booktabs}
\usepackage{multirow}
\usepackage{graphicx}

\usepackage[normalem]{ulem}
\useunder{\uline}{\ul}{}

\usepackage{amssymb}
\usepackage{pifont}

\usepackage[table]{xcolor}

\definecolor{yellow}{rgb}{1, 1, 0.7}
\definecolor{tableorange}{rgb}{1, 0.85, 0.7}
\definecolor{tablered}{rgb}{1, 0.7, 0.7}

\definecolor{cvprblue}{rgb}{0.21,0.49,0.74}
\usepackage[pagebackref,breaklinks,colorlinks,allcolors=cvprblue]{hyperref}

\def\paperID{***} 
\def\confName{CVPR}
\def\confYear{2026}

\title{SCE-SLAM: Scale-Consistent Monocular SLAM via Scene Coordinate Embeddings}

\author{
Yuchen Wu$^1$ \quad Jiahe Li$^1$ \quad Xiaohan Yu$^2$ \quad Lina Yu$^3$   \quad Jin Zheng$^1$ \quad Xiao Bai$^{1*}$\\
$^1$School of Computer Science and Engineering, State\, Key\, Laboratory\,of\,Complex\,\&\,Critical \\Software Environment,\,Jiangxi Research Institute,\,Beihang University\\
$^2$Macquarie University \\ 
$^3$Beijing Key Laboratory of Semiconductor Neural Network Intelligent Sensing\\ and Computing Technology
}

\begin{document}
\maketitle

\begin{abstract}
Monocular visual SLAM enables 3D reconstruction from internet video and autonomous navigation on resource-constrained platforms, yet suffers from scale drift, i.e., the gradual divergence of estimated scale over long sequences. Existing frame-to-frame methods achieve real-time performance through local optimization but accumulate scale drift due to the lack of global constraints among independent windows. To address this, we propose SCE-SLAM, an end-to-end SLAM system that maintains scale consistency through scene coordinate embeddings, which are learned patch-level representations encoding 3D geometric relationships under a canonical scale reference. The framework consists of two key modules: geometry-guided aggregation that leverages 3D spatial proximity to propagate scale information from historical observations through geometry-modulated attention, and scene coordinate bundle adjustment that anchors current estimates to the reference scale through explicit 3D coordinate constraints decoded from the scene coordinate embeddings. Experiments on KITTI, Waymo, and vKITTI demonstrate substantial improvements: our method reduces absolute trajectory error by 8.36m on KITTI compared to the best prior approach, while maintaining 36 FPS and achieving scale consistency across large-scale scenes.
\end{abstract}    
\section{Introduction}
\label{sec:intro}

Monocular visual SLAM has become foundational for mobile autonomous systems and internet-scale 3D reconstruction~\cite{vipe,megasam}. Recent frame-to-frame methods~\cite{teed2023dpvo,teed2021droid} have achieved impressive real-time performance through matching-based optimization, making them attractive for resource-constrained deployment. However, monocular visual SLAM faces a fundamental challenge: \textit{scale ambiguity}. Unlike stereo or RGB-D systems~\cite{kinectfusion,keetha2024splatam,CG-SLAM,niceslam} with inherent metric information, monocular cameras can only recover geometry up to an unknown scale factor.

\begin{figure}[t]
\centering
\vspace{5mm}
\includegraphics[width=\linewidth]{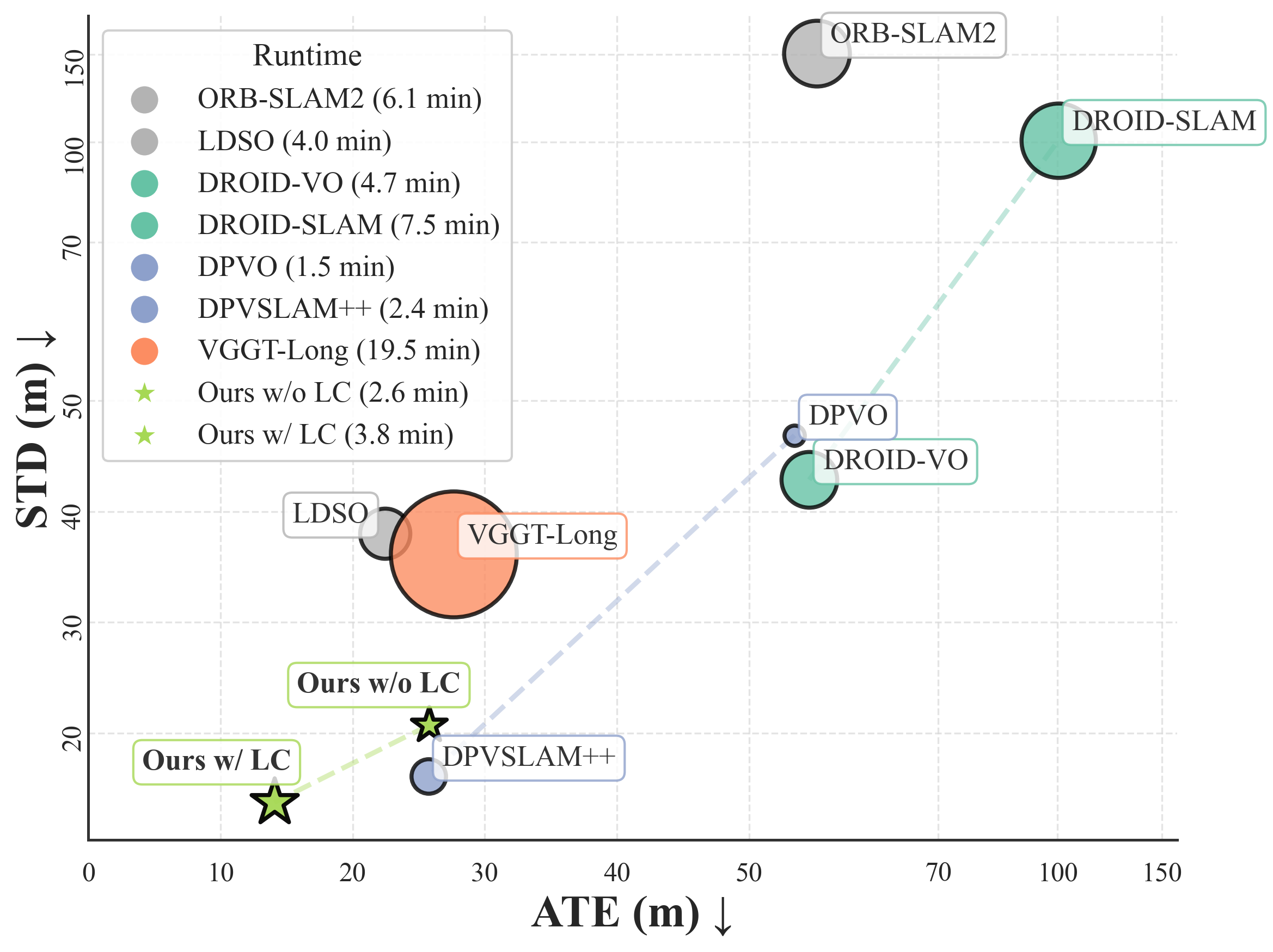}

\caption{\textbf{Kilometer-Scale Sequences Performance.} Remarkably relieving scale drift problem, our method, even without loop closure, achieves superior accuracy and robustness with high efficiency, obtaining the best ATE in average and std dev. across 11 KITTI~\cite{geiger2012kitti} scenes. Marker size denotes runtime on sequence 05.}

\label{fig:teaser}
\end{figure}

The core issue manifests as \textit{scale drift}, the progressive loss of scale consistency over long trajectories. Frame-to-frame methods optimize geometry within local sliding windows using pixel-level matching constraints that are inherently scale-agnostic: a scene scaled by any constant factor produces identical correspondences. As windows slide forward without explicit scale anchoring, each optimizes independently under potentially different scales, causing severe map fragmentation over thousands of frames that degrades loop closure and prevents reliable long-term mapping.

Recent efforts address scale drift through metric depth prediction methods~\cite{yin2023metric3d,hu2024metric3dv2, yang2024depthanything,piccinelli2024unidepth}, large multi-view geometry models~\cite{leroy2024grounding, wang2025vggt,murai2025mast3r, deng2025vggt,wang2025pi3}, and frame-to-model approaches~\cite{zhu2022nice,keetha2024splatam,deng2025gigaslam,zhang2024hi2}. However, these methods impose significant computational overhead that limits real-time applicability. This motivates our central question: \textit{Can we achieve scale-consistent monocular SLAM while preserving the efficiency of frame-to-frame optimization?}

Our key insight is that scale drift stems from the absence of temporal scale memory, where each sliding window independently establishes its own scale reference without retaining the scales of previous windows. Unlike optical flow, which captures instantaneous pixel displacements, scale represents a persistent geometric invariant of the environment that should remain constant throughout a sequence. Traditional bundle adjustment lacks explicit mechanisms to maintain this invariance across windows. To address this, we present \textit{\textbf{S}cene-\textbf{C}oordinate-\textbf{E}mbedding \textbf{SLAM}} (SCE-SLAM), an end-to-end SLAM system that maintains scale consistency through learned patch-level representations encoding 3D geometric relationships under a canonical scale reference. These representations accumulate scale-consistent information across time through recurrent updates, forming a persistent geometric memory that prevents drift across temporal windows.

We instantiate this system through a dual-branch architecture extending DPVO~\cite{teed2023dpvo}: a flow branch for pixel-level constraints, and a scene coordinate branch maintaining global scale consistency through two synergistic modules. 

\textit{Geometry-Guided Scale Propagation} updates each patch's embedding by selectively aggregating scale information from spatially nearby historical patches via geometry-modulated attention, enabling efficient accumulation of scale-consistent information across temporal windows while maintaining frame-level coordination.

\textit{Scene Coordinate Bundle Adjustment} decodes embeddings into scale-anchored 3D coordinate predictions $\mathbf{X}^{\text{prior}}$ and augments standard reprojection-based optimization with explicit coordinate constraints, actively pulling drifting estimates to the canonical scale reference. Together, these modules form a feedback loop that continuously reinforces scale consistency without requiring global optimization.

Unlike prior methods relying on external metric depth priors or heavy global representations, our system internalizes scale information within lightweight patch-level embeddings, maintaining consistency rather than predicting absolute metric scale. This enables real-time performance comparable to frame-to-frame methods while achieving scale stability approaching global optimization.

Our contributions are summarized as:
\begin{itemize}[leftmargin=*,noitemsep,topsep=2pt]
\item We introduce SCE-SLAM, an end-to-end SLAM system that maintains scale consistency through learned patch-level representations encoding geometric relationships under a canonical scale reference.

\item We propose two synergistic modules: \textit{Geometry-Guided Scale Propagation} that propagates scale information from spatially nearby historical patches via geometry-modulated attention, and \textit{Scene Coordinate Bundle Adjustment} that anchors current estimates to the reference scale through explicit 3D coordinate constraints decoded from learned embeddings.

\item We demonstrate substantial improvements across multiple benchmarks. Our method achieves an 8.36m reduction in absolute trajectory error on KITTI compared to the best prior approach, while maintaining real-time performance (36 FPS) and exhibiting superior scale consistency across extended sequences where prior frame-to-frame methods exhibit severe scale fragmentation.
\end{itemize}

\section{Related Work}
\label{sec:related}

\subsection{Monocular Visual SLAM}
Early monocular SLAM systems~\cite{mur2015orb,mur2017orbslam2,campos2021orb} pioneered the use of feature matching and bundle adjustment for camera tracking and mapping. However, these methods require careful initialization to establish scale and suffer from drift over long sequences. While effective for small-scale environments, classical methods struggle with scale consistency in extended trajectories due to their reliance on local optimization windows.
Recent advances in deep learning have enabled end-to-end approaches. TartanVO~\cite{wang2021tartanvo} explores learned optical flow for visual odometry but similarly lacks scale awareness. DROID-SLAM~\cite{teed2021droid} leverages correlation volumes and dense bundle adjustment, demonstrating impressive robustness to challenging conditions. DPVO~\cite{teed2023dpvo} further improves efficiency through patch-based optimization and recurrent updates, achieving real-time performance while maintaining accuracy. These methods represent the state-of-the-art in learning-based visual odometry, offering robustness and real-time performance. However, their optical flow constraints are inherently scale-agnostic, which is a fundamental limitation that affects all frame-to-frame optimization approaches. Our method addresses this by introducing scale-consistent geometric constraints across temporal windows while maintaining computational efficiency comparable to recent methods.

\subsection{Scale Estimation for Monocular Systems}

The success of monocular depth estimation~\cite{yang2024depthanything,birkl2023midas,bhat2023zoedepth,piccinelli2024unidepth,yin2023metric3d} has inspired attempts to leverage these models for scale supervision in SLAM. However, directly applying these models to SLAM faces two critical challenges. First, frame-by-frame predictions lack temporal consistency. Consecutive frames of the same scene may be predicted at different scales despite unchanged geometry. Second, these heavy models cannot be efficiently integrated into iterative bundle adjustment loops.
More recently, remarkable works explored learning-based multi-view geometry in a general transformer model \cite{wang2024dust3r, murai2025mast3r, wang2025vggt,wang2025cut3r}. MAST3R-SLAM~\cite{murai2025mast3r} and VGGT-Long~\cite{deng2025vggt} apply these ideas to SLAM. While promising, these methods face scalability challenges: all-to-all attention incurs quadratic complexity, limiting the number of frames that can be jointly processed. When applied to sequential windows in SLAM, these methods face two limitations: quadratic attention complexity limits the number of jointly processable frames, and independent window predictions converge to inconsistent scales. In contrast, our method maintains scale consistency through iterative geometric reasoning over local temporal windows, avoiding both the quadratic complexity of global attention and the scale drift from independent window processing.

\begin{figure*}[!t]
\centering
\includegraphics[width=\linewidth]{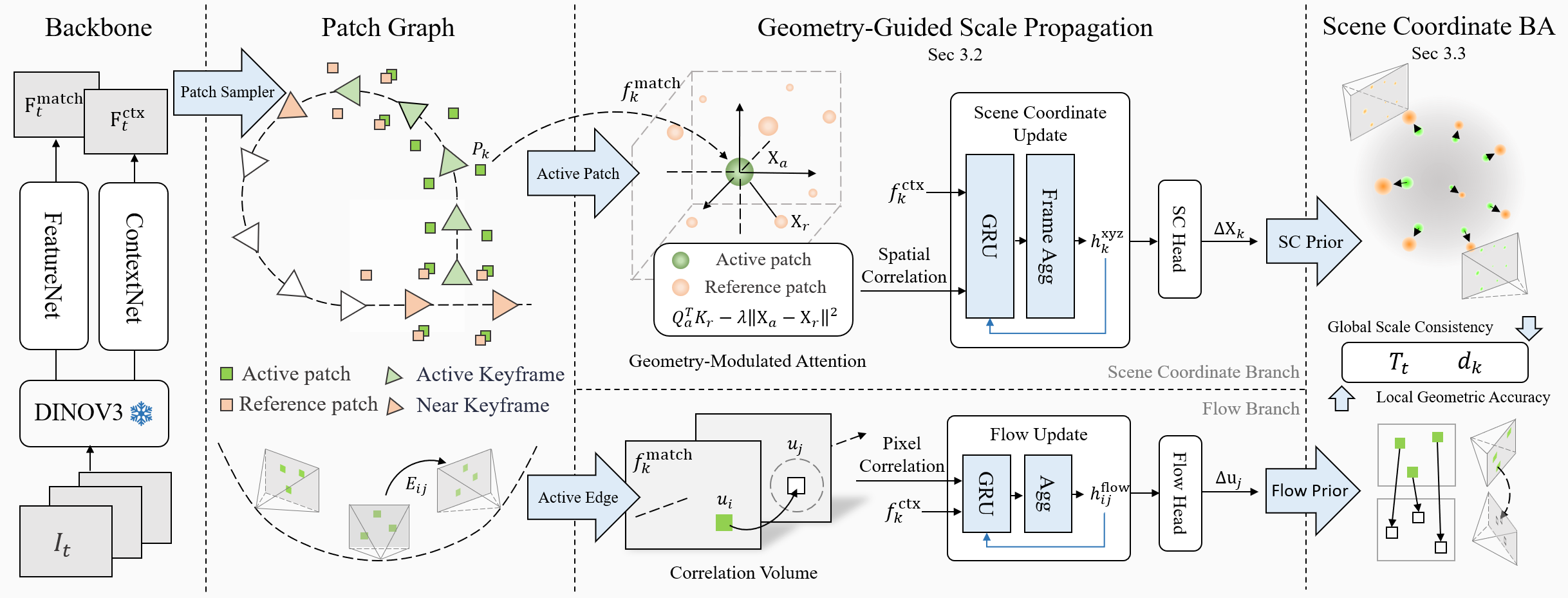}

\caption{\textbf{Overview of SCE-SLAM.} Given patch features from a DINOv3-augmented backbone, we construct a patch graph distinguishing \textit{active patches} (current window) from \textit{reference patches} (scale anchors). \textbf{Geometry-Guided Scale Propagation} operates through dual branches: the flow branch initializes correspondences and scale reference; the scene coordinate branch maintains embeddings $\mathbf{h}^{\text{xyz}}$, aggregates scale from reference patches via geometry-modulated attention to compute spatial correlation $\mathbf{f}_k^{\text{sc}}$, and predicts metric increments $\Delta \mathbf{X}_k$. \textbf{Scene Coordinate Bundle Adjustment} jointly refines poses and depths using both predictions for scale-consistent SLAM.}
\label{fig:framework}
\end{figure*}

\subsection{Learnable Scene Representations}

Several previous SLAMs \cite{sucar2021imap, zhu2022nice, yang2022vox} built upon Neural Radiance Fields (NeRF)~\cite{mildenhall2021nerf, muller2022instant} to leverage the power of implicit coordinate-based representations. While effective for capturing scene geometry, these methods require expensive volumetric rendering for pose optimization. 3D Gaussian Splatting~\cite{kerbl20233d} offers faster rendering but still incurs significant overhead for SLAM applications~\cite{keetha2024splatam,matsuki2024monogs,zhang2023hi,homeyer2025droid, wu2025vings, deng2025gigaslam}. Our approach avoids explicit scene representations, instead learning lightweight patch-level embeddings that encode scale-consistent geometry through iterative refinement and recurrent updates across temporal windows.

Scene coordinate regressions  methods\cite{brachmann2017dsac,brachmann2023ace,wang2024glace,jiang2025rscore,brachmann2024ace0} learn to predict 3D coordinates for camera relocalization. While our embeddings serve a different purpose (scale consistency rather than relocalization), we draw inspiration from their approach of encoding 3D geometric information in learned features. However, unlike these scene-specific methods, our embeddings capture generic scale relationships that generalize across arbitrary sequences.

\section{Method}
\label{sec:method}

Frame-to-frame visual SLAM optimizes geometry within sliding windows using matching constraints. While efficient, matching provides only pixel-level correspondences without scale information. As windows advance without global anchoring, scale drift accumulates over time, leading to map fragmentation.

We present SCE-SLAM, which maintains scale consistency through scene coordinate embeddings $\mathbf{h}^{\text{xyz}}$ that encode geometric relationships under a canonical scale reference. The key insight is that scale consistency can be maintained through two synergistic mechanisms: (1) \textit{Geometry-Guided Scale Propagation} (\ref{sec:aggregation}) that updates embeddings by aggregating scale information from spatially nearby historical patches, and (2) \textit{Scene Coordinate Bundle Adjustment} (\ref{sec:scba}) that anchors current estimates to the reference scale through explicit 3D coordinate constraints. We instantiate this system through a dual-branch architecture extending DPVO~\cite{teed2023dpvo} (Figure~\ref{fig:framework}), where an optical flow branch provides pixel-level constraints for tracking while a scene coordinate branch implements the above mechanisms to maintain scale consistency globally.

\subsection{Scale-Embedded Dual-Branch Architecture}
\label{sec:architecture}
\textbf{Feature Extraction and Patch Sampling.}
For each input frame $I_t$, we extract matching features $\mathbf{F}^{\text{match}}_t$ using FeatureNet and context features $\mathbf{F}^{\text{ctx}}_t$ using ContextNet. Each encoder consists of a lightweight CNN backbone augmented with pre-trained DINOv3~\cite{simeoni2025dinov3} features. Specifically, we extract DINOv3 features at the same spatial resolution and fuse them with CNN features via a $1\times1$ convolution. This design combines the generalization capability of foundation models with task-specific learned representations, providing robust feature matching across diverse scenes while maintaining computational efficiency.

Unlike DPVO's random sampling, which suffices for local optical flow prediction, scale consistency learning requires patches that maintain stable correspondences across multiple views. Random sampling produces patches that may only be visible in one or two frames, limiting the network's ability to learn multi-view geometric constraints. Instead, we extract 80 patches per frame based on SuperPoint~\cite{detone2018superpoint} keypoints. Following DPVO, we extract $3 \times 3$ patches from both feature maps, yielding $\mathbf{f}_k^{\text{match}}, \mathbf{f}_k^{\text{ctx}}$ for patch $p_k$. This keypoint-based sampling ensures: (1) patches correspond to trackable visual features, and (2) multi-view observability that is essential for scale consistency learning. These features serve as the foundation for our dual representation of geometry, where we maintain two hidden states.

\noindent\textbf{Complementary Hidden State Representations.} For each patch $p_k$, we maintain two types of hidden states that capture different aspects of geometry:

1) \emph{Edge-based flow states} $\mathbf{h}^{\text{flow}} \in \mathbb{R}^{384}$ (inherited from DPVO): These encode pixel-level motion information for each edge $(i,j)$. Updated using correlation volumes computed from matching features across frames, these states drive precise optical flow prediction and provide scale-agnostic geometric constraints.

2) \emph{Scene coordinate embeddings} $\mathbf{h}^{\text{xyz}} \in \mathbb{R}^{384}$: These are the core learned representations of our system, encoding each patch's geometric context under the canonical scale reference. Critically, these embeddings are patch-centric rather than edge-centric, allowing them to accumulate scale-consistent information from all frames observing the patch. Initialized to zero, they are updated through Geometry-Guided Scale Propagation (detailed in \ref{sec:aggregation}).

The distinction is fundamental: $\mathbf{h}^{\text{flow}}$ captures instantaneous pairwise motion (scale-agnostic), while $\mathbf{h}^{\text{xyz}}$ accumulates persistent geometric relationships (scale-aware). This dual representation enables our system to leverage both local pixel-level accuracy and global scale consistency.

\noindent\textbf{Dual Prediction Heads.}The flow branch predicts 2D optical flow increments and predicted confidence weights for each edge $(i,j)$, following DPVO's formulation. The scene coordinate branch decodes embeddings into 3D coordinate predictions for each patch $k$ in frame $t$:
\begin{equation}
    \Delta \mathbf{X}_k, \, w_k = \text{SCHead}(\mathbf{h}_k^{\text{xyz}}), \quad
    \mathbf{X}_k^{\text{prior}} = \mathbf{X}_k + \Delta \mathbf{X}_k
    \label{eq:coord_prediction}
\end{equation}
Where $\mathbf{X}_k = \mathbf{T}_t \cdot \pi^{-1}(\mathbf{u}_k, d_k)$ transforms the patch to world coordinates using the current iteration's depth/pose estimates. Rather than directly predicting absolute coordinates (which would discard geometric information from bundle adjustment), the network predicts residuals $\Delta \mathbf{X}_k$ that refine current estimates toward the scale-consistent reference encoded in the embeddings. The resulting $\mathbf{X}_k^{\text{prior}}$ represents the patch's expected spatial position under the canonical scale reference, serving as an anchoring constraint in the next optimization iteration (\ref{sec:scba}). The confidence weight $w_k$ directly modulates the strength of this constraint based on the embedding's certainty.

\subsection{Geometry-Guided Scale Propagation}
\label{sec:aggregation}

The first key module maintains scale consistency across temporal windows by updating scene coordinate embeddings $\mathbf{h}^{\text{xyz}}$ through selective integration of historical geometric observations. The core challenge is: how can current patches leverage the canonical scale encoded in historical embeddings without incurring prohibitive computational cost or introducing spurious correlations? A naive approach would apply global self-attention across all patches, allowing each current patch to attend to all historical patches. However, this suffers from two critical limitations.

1) \emph{Computational prohibitiveness.} As patches accumulate across the sequence, full attention between all current and historical patches grows quadratically with sequence length. For trajectories spanning thousands of frames with tens of thousands of accumulated patches, this rapidly becomes computationally intractable.

2) \emph{Geometric implausibility.} Not all patches should interact equally. A patch observing a nearby wall should not attend to one observing a distant mountain, as such spurious long-range dependencies degrade geometric reasoning. Moreover, patches with poor geometric estimates (indicated by high bundle adjustment residuals) should not serve as reliable scale references.

Our solution exploits a key insight: \textit{scale consistency propagates through spatial proximity in 3D space}. If two patches observe physically nearby 3D points, their geometric relationship under the canonical scale should be consistent. We thus design a four-stage propagation mechanism.

\noindent\textbf{Reference Patch Selection.} Before aggregation, we construct a reliable reference set $\mathcal{R}$ from historical patches. We rank all historical patches by their bundle adjustment residuals from previous iterations and retain only the top 50\% with the lowest residuals, ensuring that scale references come from geometrically consistent observations. Additionally, to bound computational cost, we temporally constrain $\mathcal{R}$ to patches from the 30 frames closest to the current sliding window's temporal center. With 80 patches extracted per frame, this adaptive selection strategy yields $\mathcal{R}=1200$ patches, making attention computation tractable for real-time operation while preserving recent, high-quality geometric information.

\noindent\textbf{Geometry-Modulated Attention.} For each active patch in the current window, we compute attention over its reference set $\mathcal{R}$ by combining feature similarity with 3D spatial proximity. We first compute attention that modulates feature-based similarity with geometric distance:
\begin{equation}
    e_{ar} = \frac{\mathbf{Q}_a^\top \mathbf{K}_r}{\sqrt{d}} 
         - \lambda \|\mathbf{X}_a - \mathbf{X}_r\|^2
    \label{eq:attn_logit}
\end{equation}
where $\mathbf{Q}_a = \mathbf{W}_q \mathbf{f}_a^{\text{match}}$ is the query from the active patch's matching features, $\mathbf{K}_r = \mathbf{W}_k \mathbf{f}_r^{\text{match}}$ and $\mathbf{V}_r = \mathbf{W}_v \mathbf{f}_r^{\text{match}}$ are keys and values from reference patch $r \in \mathcal{R}$, $\mathbf{X}_a$ is the current patch's 3D position from the latest optimization, and $\mathbf{X}_r$ is the reference patch's stored 3D position. The learnable parameter $\lambda$ balances feature and geometric cues. The geometric penalty $-\lambda \|\mathbf{X}_a - \mathbf{X}_r\|^2$ progressively down-weights attention to spatially distant patches, encoding the natural inductive bias that nearby patches in 3D should share consistent scale understanding.

To further enhance geometric awareness, we encode relative 3D displacements with MLP and directly fuse them with the value vectors:
\begin{equation}
    \mathbf{V}_r^{\text{geo}} = \mathbf{V}_r + \text{MLP}_{\text{pos}}\left(\mathbf{X}_a - \mathbf{X}_r
    \right)
    \label{eq:geo_value}
\end{equation}
where $\text{MLP}_{\text{pos}}$ encodes the displacement. This allows the network to reason effectively about not just whether patches are nearby (via the attention penalty), but also their precise relative spatial configuration.

We then aggregate information from the reference set by computing weighted combinations of the geometric value vectors, yielding the \textit{spatial correlation} for patch $a$:
\begin{equation}
    \mathbf{f}^{\text{sc}} = \text{MLP}_{\text{sc}}\left(
         e_{ar} \cdot \mathbf{V}_r^{\text{geo}}
    \right)
    \label{eq:spatial_correlation}
\end{equation}
where the attention $e_{ar}$ directly weight the contributions from reference patches, and $\text{MLP}_{\text{sc}}$ projects the aggregated features to match the dimensionality of $\mathbf{h}^{\text{xyz}}$. This aggregated spatial correlation $\mathbf{f}^{\text{sc}}$ encapsulates scale-consistent geometric information from spatially coherent historical observations, analogous to learned pixel correlation in optical flow estimation but operating in 3D space. When the current window begins to drift in scale, this aggregation mechanism anchors patches to their spatial neighbors in $\mathcal{R}$, which encode the canonical scale reference established during initialization, effectively pulling drifting estimates back toward scale consistency.

\begin{table*}[!t]
\resizebox{\textwidth}{!}{%
\begin{tabular}{@{}l|lc|cc|ccccccccccc}
\toprule
 & \textbf{Methods} & \textbf{LC} & \textbf{Avg.} & \textbf{Std.} & \textbf{00} & \textbf{01} & \textbf{02} & \textbf{03} & \textbf{04} & \textbf{05} & \textbf{06} & \textbf{07} & \textbf{08} & \textbf{09} & \textbf{10} \\ \cmidrule(l){2-16} 
 & ORB-SLAM2 & \checkmark & 54.82 & 150.79 & \textbf{6.03} & 508.34 & \textbf{14.76} & \textbf{1.02} & 1.57 & \textbf{4.04} & \textbf{11.16} & 2.19 & \textbf{38.85} & \textbf{8.39} & \textbf{6.63} \\
\multirow{-3}{*}{{\rotatebox[origin=c]{90}{\small Classic}}} & LDSO & \checkmark & {\ul 22.43} & 38.04 & 9.32 & 11.68 & 31.98 & 2.85 & 1.22 & {\ul 5.10} & 13.55 & 2.96 & 129.02 & {\ul 21.64} & 17.36 \\ \midrule
 & DROID-VO & \ding{55} & 54.19 & 42.86 & 98.43 & 84.2 & 108.8 & 2.58 & 0.93 & 59.27 & 64.40 & 24.20 & 64.55 & 71.80 & 16.91 \\
 & DPVO & \ding{55} & 53.61 & 46.87 & 113.21 & 12.69 & 123.4 & 2.09 & 0.68 & 58.96 & 54.78 & 19.26 & 115.90 & 75.10 & 13.63 \\
 & DROID-SLAM & - & 100.28 & 100.53 & 92.10 & 344.6 & 107.61 & 2.38 & 1.00 & 118.5 & 62.47 & 21.78 & 161.6 & 72.32 & 118.7 \\
 & DPV-SLAM & \checkmark & 53.03 & 48.60 & 112.80 & 11.50 & 123.53 & 2.50 & 0.81 & 57.80 & 54.86 & 18.77 & 110.49 & 76.66 & 13.65 \\
 & DPV-SLAM++ & \checkmark & 25.75 & {\ul 16.11} & 8.30 & 11.86 & 39.64 & 2.50 & 0.78 & 5.74 & 11.60 & \textbf{1.52} & 110.90 & 76.70 & 13.70 \\ \cmidrule(l){2-16} 
 & MASt3R-SLAM & - & / & / & TL & TL & TL & TL & TL & TL & TL & TL & TL & TL & TL \\
 & CUT3R & - & / & / & OOM & OOM & OOM & 148.07 & 22.31 & OOM & OOM & OOM & OOM & OOM & OOM \\
 & VGGT-Long & \checkmark & 27.64 & 36.15 & 8.67 & 121.17 & 32.08 & 6.12 & 4.23 & 8.31 & {\ul 5.34} & 4.63 & 53.10 & 41.99 & 18.37 \\ \cmidrule(l){2-16} 
 & \textbf{Ours} & \ding{55} & {\color[HTML]{009901} 25.79} & {\color[HTML]{009901} 20.73} & {\color[HTML]{009901} 53.31} & {\color[HTML]{008801} \textbf{9.63}} & {\color[HTML]{009901} 62.32} & {\color[HTML]{009901} {\ul 2.05}} & {\color[HTML]{008801} \textbf{0.54}} & {\color[HTML]{009901} 31.34} & {\color[HTML]{009901} 26.37} & {\color[HTML]{009901} 11.65} & {\color[HTML]{009901} {\ul 43.67}} & {\color[HTML]{009901} 30.19} & {\color[HTML]{009901} {\ul 12.57}} \\
\multirow{-10}{*}{{\rotatebox[origin=c]{90}{\small Learning-based}}} & \textbf{Ours} & \checkmark & \textbf{14.07} & \textbf{13.76} & {\ul 8.11} & {\ul 9.82} & {\ul 31.04} & 2.40 & {\ul 0.56} & 5.40 & 11.17 & {\ul 1.95} & 41.21 & 30.38 & 12.66 \\ \bottomrule
\end{tabular}%
}
\caption{\textbf{Comparison on the KITTI Dataset of ATE RMSE $\downarrow$ (m).} The \textbf{best} and {\ul second-best} results are marked. The best results from methods without loop closure (LC) are in green. Our method achieves the best in the average error (Avg.) and standard deviation (Std.) compared to the methods with or without loop closure, respectively, demonstrating the superior accuracy and robustness.}
\label{tab:kitti}
\end{table*}

\begin{table*}[]
\resizebox{\textwidth}{!}{
\begin{tabular}{l|cc|ccccccccc}
\toprule
\textbf{Methods}     & \textbf{Avg.}           & \textbf{Std.}           & \textbf{163453191}     & \textbf{183829460}       & \textbf{315615587}      & \textbf{346181117}      & \textbf{371159869}   & \textbf{405841035}      & \textbf{460417311}      & \textbf{520018670}      & \textbf{610454533}      \\ \midrule
DROID-SLAM  & 4.396          & 3.829          & 3.705         & 0.301           & 0.447          & 8.653          & 9.320       & 7.621          & 4.170          & TL             & 0.264          \\
DPV-SLAM++  & 3.874          & 8.491          & 1.942         & {\ul 0.065}     & 0.181          & {\ul 0.269}    & 4.980       & 0.969          & {\ul 0.177}    & 26.118         & 0.168          \\
MASt3R-SLAM & 5.560          & 4.120          & 4.500           & 0.556           & 1.833          & 12.544         & 8.601       & 1.412          & 5.428          & 7.910          & 1.195          \\
MegaSaM     & 2.776          & 6.220          & 0.556        & \textbf{0.043} & \textbf{0.112} & 2.288          & 19.259      & \textbf{0.651} & 1.143          & \textbf{0.808} & \textbf{0.123} \\
VGGT-Long   & {\ul 1.996}    & {\ul 1.098}    & {\ul 1.753}   & 2.629           & 0.559          & 3.452          & 3.343       & 1.444          & 1.541          & 2.547          & 0.455          \\
\textbf{Ours}        & \textbf{0.915} & \textbf{1.049} & \textbf{1.620} & 0.076           & {\ul 0.118}    & \textbf{0.219} & {\ul 2.813} & {\ul 0.783}    & \textbf{0.176} & {\ul 2.248}    & {\ul 0.184}    \\ \bottomrule
\end{tabular}
}
\caption{\textbf{Comparison on the Waymo Dataset of ATE RMSE $\downarrow$ (m).} The \textbf{best} and {\ul second-best} results are marked. Our method achieves the best performance in the average error (Avg.) and standard deviation (Std.) compared to the SOTA methods.}
\label{tab:waymo}
\end{table*}

\noindent\textbf{Embedding Update via Recurrent Aggregation.} The spatial correlation is then combined with the patch's own context features and the previous embedding state through residual connections:
\begin{equation}
    \tilde{\mathbf{h}}^{\text{xyz}} = \mathbf{h}^{\text{xyz}} 
        + \mathbf{f}^{\text{sc}} + \mathbf{f}^{\text{ctx}}
    \label{eq:hidden_aggregate}
\end{equation}
This additive fusion preserves existing geometric memory in $\mathbf{h}^{\text{xyz}}$ while integrating new scale information from spatial neighbors ($\mathbf{f}^{\text{sc}}$) and current visual observations ($\mathbf{f}^{\text{ctx}}$). We then pass the updated embeddings through a GRU cell to enable temporal information flow across iterations:
\begin{equation}
    \mathbf{h}^{\text{xyz}} = \text{GRU}\left(\tilde{\mathbf{h}}^{\text{xyz}}\right)
    \label{eq:hidden_gru}
\end{equation}
The GRU's recurrent structure allows $\mathbf{h}^{\text{xyz}}$ to accumulate geometric memory across multiple iterations, essential for maintaining the scale reference over long sequences spanning thousands of frames.

\noindent\textbf{Frame-Level Coordination.} Finally, we exploit the structural coupling between patches within the same frame. Since all patches in frame $t$ share the same camera pose, their scale estimates should be mutually consistent. To capture this coupling, we perform frame-level aggregation before decoding scene coordinates:
\begin{equation}
    {\mathbf{h}}^{\text{xyz}} = \text{FrameAgg}\left(
    \mathbf{h}^{\text{xyz}}
    \right)
    \label{eq:frame_agg}
\end{equation}

\subsection{Scene Coordinate Bundle Adjustment}
\label{sec:scba}

The second key module translates learned embeddings into explicit optimization constraints that anchor scale across temporal windows. We formulate bundle adjustment as jointly optimizing camera poses $\mathbf{T}_t$ and patch depths $d_k$ over two complementary objectives via standard Gauss-Newton optimization.

\noindent\textbf{Dual Optimization Objectives.} Our bundle adjustment balances two complementary residual terms that serve distinct geometric roles:

\textit{Flow-based residuals} $\mathbf{r}_{ij}^{\text{flow}}$ enforce local geometric accuracy through pixel-level correspondences:
\begin{equation}
    \mathbf{r}_{ij}^{\text{flow}} = w_{ij}^{\text{flow}} 
    \left(\mathbf{u}_j^{\text{prior}} - \pi(\mathbf{T}_j \cdot \mathbf{T}_i^{-1} 
    \cdot \pi^{-1}(\mathbf{u}_i, d_i))\right)
    \label{eq:reproj_residual}
\end{equation}
for each edge $(i,j) \in \mathcal{E}$ (frame pairs with optical flow predictions), where $\mathbf{u}_j^{\text{prior}}$ is the flow-predicted target location, and $w_{ij}^{\text{flow}}$ is the flow confidence weight. These pixel-space residuals drive high-fidelity geometric reconstruction within the sliding window but are inherently scale-agnostic, as any uniform scaling of depths and baselines produces identical pixel correspondences.

\textit{Scene coordinate residuals} $\mathbf{r}_k^{\text{SC}}$ enforce global scale consistency through 3D coordinate constraints:
\begin{equation}
    \mathbf{r}_k^{\text{xyz}} = w_k^{\text{xyz}} 
    \left(\mathbf{X}_k^{\text{prior}} - \mathbf{T}_{t(k)} \cdot \pi^{-1}(\mathbf{u}_k, d_k)\right)
    \label{eq:sc_residual}
\end{equation}
for each patch $k$, where $\mathbf{X}_k^{\text{prior}}$ is the scale-anchored coordinate prediction decoded from embedding $\mathbf{h}_k^{\text{xyz}}$ (Eq.~\ref{eq:coord_prediction}), $\mathbf{T}_{t(k)}$ is the pose of patch $p_k$'s source frame, and $w_k^{\text{xyz}}$ weights the constraint by prediction confidence. Unlike flow residuals, these operate in world coordinates and explicitly penalize scale deviations: if current estimates drift from the canonical scale encoded in $\mathbf{X}_k^{\text{prior}}$ (established via Geometry-Guided Scale Propagation, \ref{sec:aggregation}), the residual magnitudes increase and generate corrective gradients that pull $\{d_k, \mathbf{T}_{t(k)}\}$ back toward scale-consistent values.

At each iteration, we alternate between minimizing flow residuals and coordinate residuals via Gauss-Newton, computing updates $\{\Delta \mathbf{T}_t, \Delta d_k\}$ that refine local geometry while maintaining global scale consistency.

\noindent\textbf{Two-Stage Bootstrapping Strategy.} To enable scale-consistent learning, we operate in two stages:

\textit{Stage 1 (Initialization):} We minimize flow residuals $\{\mathbf{r}_{ij}^{\text{flow}}\}$ to obtain initial estimates $\{\mathbf{T}_t, d_k\}$ that establish the canonical scale reference for the sequence.

\textit{Stage 2 (Scale-anchored refinement):} We update embeddings 
$\mathbf{h}_k^{\text{xyz}}$ via Geometry-Guided Scale Propagation 
(\ref{sec:aggregation}) and decode them into coordinate priors 
$\mathbf{X}_k^{\text{prior}}$ (Eq.~\ref{eq:coord_prediction}). Within each iteration, we perform one round of BA with scene coordinate residuals to anchor scale, followed by two rounds with only flow residuals to refine local geometry.

This two-stage design bootstraps scale-consistent predictions from flow-based initialization, enabling the coordinate branch to internalize and propagate the reference scale throughout long sequences.

\begin{figure*}[!t]
\centering
\includegraphics[width=1\linewidth]{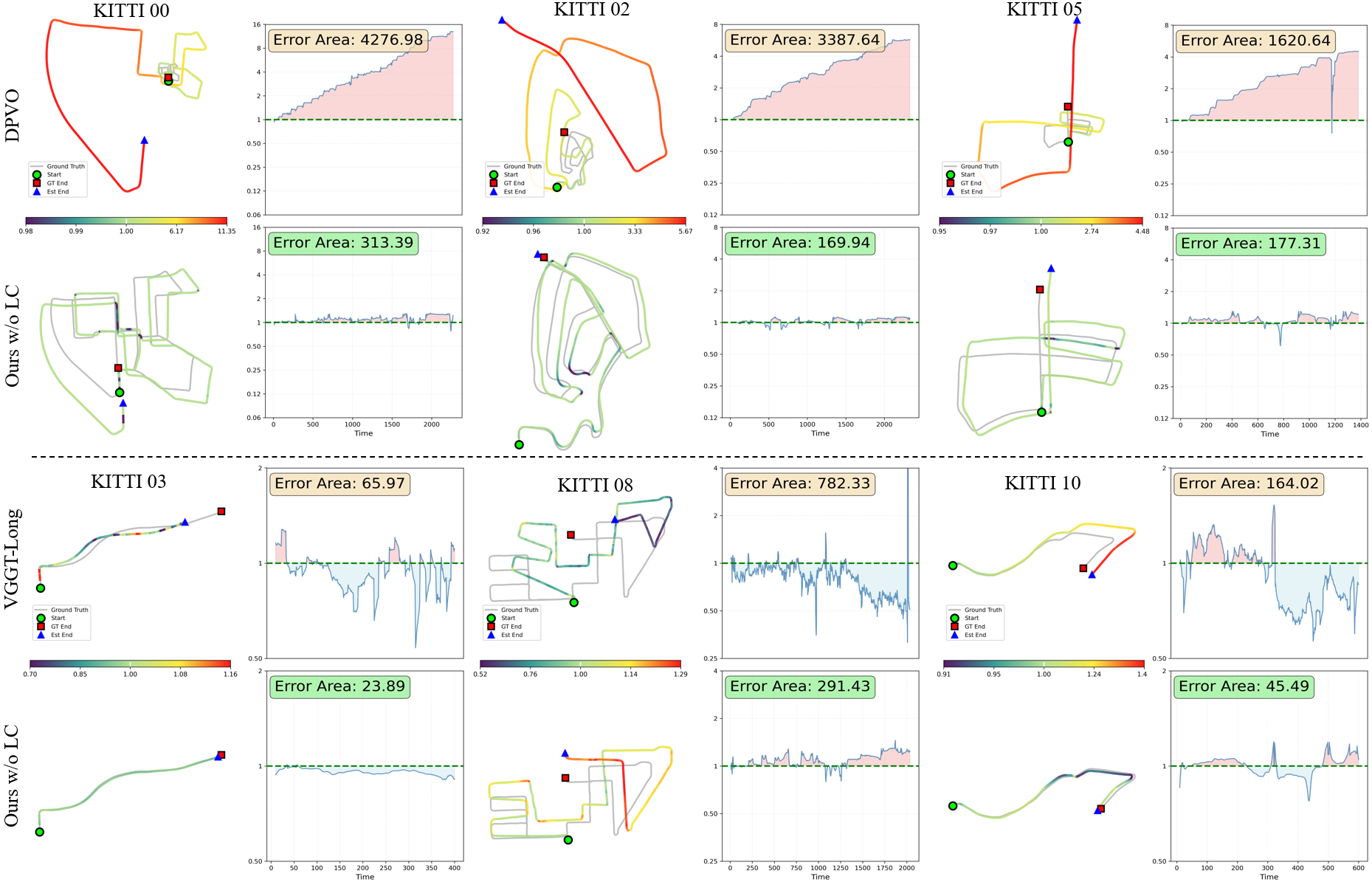}
\caption{\textbf{Visual Odometry Comparison on the KITTI Datasets w/o Loop Closure.} We visualize the estimated trajectory with the scale from the first 20 timestamps. The results show the strong VO capability of our method. Especially, we visualize the estimated relative scale by color (left), and logarithmic bias per time (right). The reduced scale drift demonstrates our advantage in the global scale consistency. }
\label{fig:comp}
\end{figure*}

\section{Experiments}

\begin{table}[]
\resizebox{\linewidth}{!}{
\begin{tabular}{l|cccccc|c}
\toprule
{\color[HTML]{656565} \textit{Condition}} & {\color[HTML]{656565} \textit{Clone}} & {\color[HTML]{656565} \textit{Fog}} & {\color[HTML]{656565} \textit{Morning}} & {\color[HTML]{656565} \textit{Overcast}} & {\color[HTML]{656565} \textit{Rain}} & {\color[HTML]{656565} \textit{Sunset}} & {\color[HTML]{656565} \textit{Avg.}}    \\ \midrule
DROID-SLAM                                                   & 1.451                                                     & 1.809                                                     & 1.339                                                     & 1.459                                                     & 1.786                                                     & 1.626                                                       & 1.578                  \\
MASt3R-SLAM                                                  & TL                                                        & TL                                                        & TL                                                        & TL                                                        & TL                                                        & TL                                                          & TL                     \\
CUT3R                                                        & 43.370                                                    & 31.627                                                    & 40.858                                                    & 39.765                                                    & 28.810                                                    & 43.684                                                      & 38.019                 \\
VGGT-Long                                                    & 2.631                                                     & 2.221                                                    & 1.932                                                     & 1.726                                                     & 2.240                                                     & 1.784                                                       & 2.089                  \\
DPV-SLAM++                                                    & {\ul 0.364}                                               & {\ul 0.498}                                              & {\ul 0.277}                                               & {\ul 0.241}                                               & {\ul 0.383}                                               & {\ul 0.299}                                                 & {\ul 0.343}            \\
\textbf{Ours}                                                & \textbf{0.291}                                            & { \textbf{0.356}}                                      & \textbf{0.240}                                            & \textbf{0.227}                                            & \textbf{0.296}                                            & \textbf{0.271}                                              & \textbf{0.280}         \\ \bottomrule
\end{tabular}
}

\caption{\textbf{Comparison on the Virtual KITTI Dataset of ATE RMSE $\downarrow$ (m).}  The \textbf{best} and {\ul second-best} results are marked.}
\label{tab:vkitti}
\end{table}

\subsection{Experimental Setup}

\noindent\textbf{Implementation.}
We train on synthetic TartanAir~\cite{tartanair} for 240K iterations using AdamW optimizer (lr=$8\times10^{-5}$, weight decay=$1\times10^{-6}$) with sequence length 15. Following DPVO, we supervise optical flow and camera poses. Additionally, for the scene coordinate branch, we supervise predictions with GT positions:
\begin{equation}
    \mathcal{L}_{\text{SC}} = \sum_{k} \left\| \mathbf{X}_k - \mathbf{X}_k^{\text{GT}} \right\|^2
\end{equation}
We combine these losses, adopting DPVO's weights for flow and pose terms, with $\lambda_1 = 0.1$ and $\lambda_2 = 10$:
\begin{equation}
    \mathcal{L}_{\text{total}} = \lambda_1 \cdot \mathcal{L}_{\text{flow}} + \lambda_2 \cdot \mathcal{L}_{\text{pose}} + \mathcal{L}_{\text{SC}}
    \label{eq:loss_total}
\end{equation}
Training follows a two-stage curriculum: the first 10K iterations freeze the flow branch to initialize the coordinate branch, then both branches are jointly optimized.

\noindent\textbf{Evaluation Datasets.} We evaluate on three large-scale benchmarks: KITTI Odometry~\cite{geiger2012kitti} (11 driving sequences), Waymo Open~\cite{Sun_2020_waymo} (9 urban sequences), and Virtual KITTI~\cite{gaidon2016virtual} (synthetic sequences under varying weather).

\noindent\textbf{Baselines.} We compare against frame-to-frame approaches (DROID-VO~\cite{teed2021droid}, DPVO~\cite{teed2023dpvo}, DPV-SLAM/++~\cite{lipson2024dpvslam}, MegaSaM~\cite{megasam}), multi-view alignment methods (MAST3R-SLAM~\cite{murai2025mast3r}, CUT3R~\cite{wang2025cut3r}, VGGT-Long~\cite{deng2025vggt}), and classical systems (ORB-SLAM2~\cite{mur2017orbslam2}, LDSO~\cite{gao2018ldso}).

\noindent\textbf{Metrics.} We report Absolute Trajectory Error (ATE) RMSE on test sequences. We also report standard deviation (STD) to measure cross-run consistency. For runtime failure cases, we mark \textbf{TL} for situations of tracking lost, and \textbf{OOM} for out of memory.

\subsection{Main Results}

\noindent\textbf{KITTI Odometry.} Table~\ref{tab:kitti} shows our method achieves the lowest average ATE (14.07m) and STD among all visual odometry methods, approaching DPV-SLAM++, which incorporates loop closure. Figure~\ref{fig:comp} visualizes scale consistency: by aligning only the first 20 frames, we preserve the scale established during initialization. Our method maintains uniform scale (consistent color) throughout the trajectory, while DPVO and VGGT-Long exhibit progressive drift, directly validating our core contribution.

Figure~\ref {fig:comp} visualizes the scale consistency advantage. Unlike conventional evaluations that align trajectories globally, we align only using the first 20 frames to preserve the scale established during initialization. Our method maintains consistent scale (uniform color) throughout long sequences, while DPVO and VGGT-Long exhibit progressive scale drift. Results validate our core contribution: maintaining scale consistency through scene coordinate embeddings prevents drift in frame-to-frame methods.

\noindent\textbf{Waymo Open Dataset.} Table~\ref{tab:waymo} demonstrates robustness to complex urban scenarios. Our method outperforms MegaSaM~\cite{megasam}, which augments tracking with metric depth from a large pre-trained model, showing that learned embeddings are more effective than external metric priors.

\noindent\textbf{Virtual KITTI.} Table~\ref{tab:vkitti} shows results across six weather 
conditions. Our method achieves the best average ATE (0.28m), surpassing even DPV-SLAM++ with its loop closure module, demonstrating consistent performance across diverse conditions.

\begin{figure}[t]
\centering
\vspace{5mm}
\includegraphics[width=\linewidth]{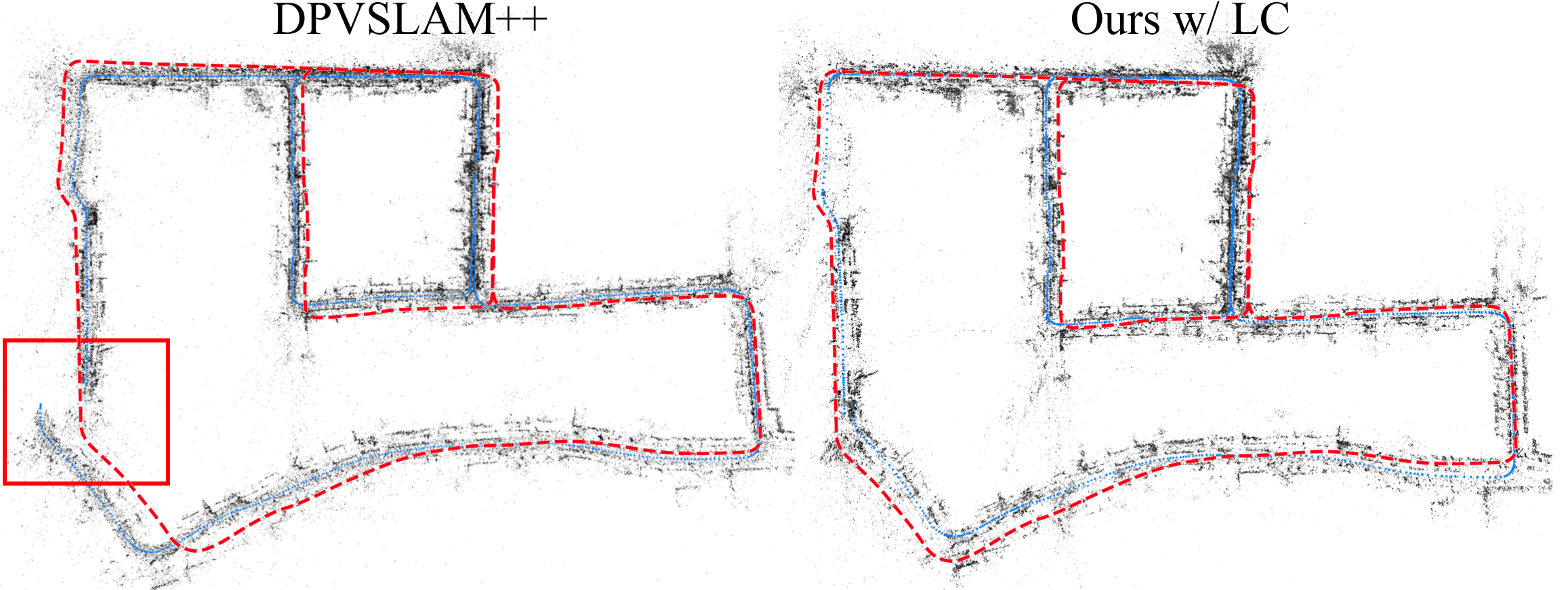}
\caption{\textbf{Trajectory on 4Seasons Neighborhood.} 
Blue: our prediction; red dashed: ground truth. Our method successfully 
closes the loop with scale consistency, while DPV-SLAM++ fails.}
\label{fig:4seasons}
\end{figure}

\noindent\textbf{Qualitative Analysis.} Beyond the scale consistency showed in Figure~\ref{fig:comp}, Figure~\ref{fig:4seasons} shows the practical impact on the challenging 4Seasons dataset~\cite{fourseasons}. Our method successfully closes the loop (blue prediction aligns with red ground truth), while DPV-SLAM++, despite having explicit loop closure modules, fails due to accumulated scale fragmentation. This validates that maintaining scale consistency is critical for reliable loop closure in long-sequence scenes.

\subsection{Ablation Study}

\noindent\textbf{Component Analysis.}
Table~\ref{tab:ablation_part} analyzes key components on KITTI (averaged over all sequences). Starting from DPVO with DINOv3 features (45.84m, row A), stronger features alone are insufficient without scale anchoring. Adding a basic coordinate branch (GRU-based $\mathbf{h}^{\text{xyz}}$ with 1200-patch aggregation) provides modest improvement (row B). Our full design, incorporating geometry-modulated attention, reference patch selection, and frame-level aggregation, reduces error to 31.83m (row C). Using SuperPoint sampling instead of random sampling further improves to 25.79m (row D), as SuperPoint ensures trackable features across views. With loop closure, we achieve 14.07m (row E).

\begin{figure}[t]
\centering
\vspace{5mm}
\includegraphics[width=\linewidth]{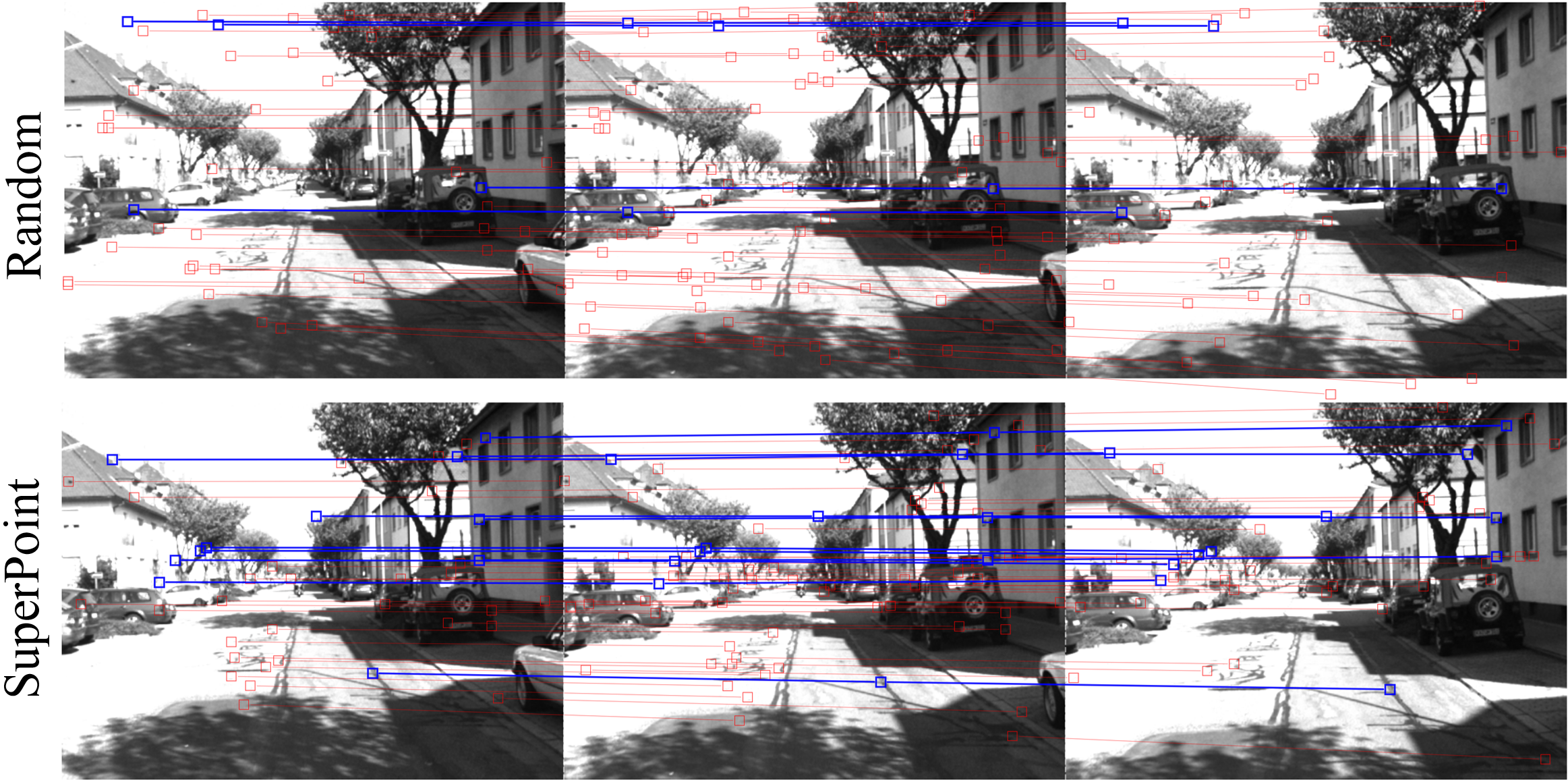}
\caption{\textbf{Multi-view patch sampling consistency.} Patch tracking 
across three consecutive frames. Blue boxes show successful tracks across 
views, red boxes show tracking failures.}
\label{fig:sample}
\end{figure}

\noindent\textbf{Cross-View Patch Consistency.}
Figure~\ref{fig:sample} compares patch sampling strategies across three frames. Random sampling produces only 4 tracks, of which 2 fall on invalid sky regions, while SuperPoint-based~\cite{detone2018superpoint} sampling yields 11 tracks, all on valid structure. This cross-view consistency is essential for the scene coordinate branch: stable multi-view observations enable reliable scale aggregation via Geometry-Guided Scale Propagation (\ref{sec:aggregation}).

\begin{table}[t]
\resizebox{\linewidth}{!}{
\setlength{\tabcolsep}{3mm}
\begin{tabular}{l|c|c|c|c}
\toprule
\textbf{Item} & \textbf{SC Brach}                    & \textbf{Sampler}    & \textbf{LC}                         & \textbf{ATE (m) $\downarrow$}   \\
\midrule
A. & \ding{55}  & Random     & \ding{55} & 45.84 \\
B. & Basic Design          & Random     & \ding{55} &   43.62    \\
C. & Geo Scale Pro & Random     & \ding{55} & 31.83 \\ \midrule
D. & Geo Scale Pro & SP & \ding{55} & 25.79 \\
E. & Geo Scale Pro & SP & \checkmark  & 14.07\\ \bottomrule
\end{tabular}
}
\caption{\textbf{Ablation study on KITTI Odometry.}}
\label{tab:ablation_part}
\end{table}

\noindent\textbf{Impact of Foundation Model Features.}
Table~\ref{tab:ablation_dino} analyzes DINOv3's~\cite{simeoni2025dinov3} role. For DPVO, DINOv3 provides a marginal improvement, as scale drift persists without anchoring mechanisms. Our method exhibits stronger synergy with DINOv3, as geometry-guided aggregation (\ref{sec:aggregation}) relies on cross-view feature matching to identify spatial neighbors. DINOv3's view-invariant representations enhance matching accuracy, yielding more reliable geometry-modulated attention weights (Eq.~\ref{eq:attn_logit}). With loop closure, the gap persists (22.91m vs. 14.07m), confirming that architectural design is critical beyond the feature.

\begin{table}[]
\resizebox{\linewidth}{!}{
\setlength{\tabcolsep}{4mm}
\begin{tabular}{l|c|c|c}
\toprule
\textbf{Method}     & \textbf{LC}  & \textbf{DINOv3} & \textbf{ATE (m) $\downarrow$}                            \\
\midrule
DPVO       &  \ding{55}   &  \ding{55}      & 53.61                          \\
DPVO       &  \ding{55}   & \checkmark    & 45.84                          \\
\textbf{Ours}       &  \ding{55}   & \checkmark    & \textbf{25.79}                          \\ \midrule
DPV-SLAM++ & \checkmark &  \ding{55}      & 25.75 \\
DPV-SLAM++ & \checkmark & \checkmark    & 22.91                          \\
\textbf{Ours}       & \checkmark & \checkmark    & \textbf{14.07}\\ \bottomrule
\end{tabular}
}
\caption{\textbf{Impact of DINOv3 features on KITTI Odometry.}}
\end{table}
\label{tab:ablation_dino}

\section{Conclusion}

We present SCE-SLAM, a monocular SLAM system that maintains scale consistency 
through learned scene coordinate embeddings. Our approach encodes geometric relationships under a canonical scale reference and propagates this information across temporal windows via geometry-modulated attention and 3D coordinate constraints in bundle adjustment. Experiments demonstrate substantial accuracy improvements while maintaining real-time efficiency.
\clearpage
\setcounter{page}{1}
\maketitlesupplementary

\renewcommand\thesection{\Alph{section}}
\setcounter{figure}{6}
\setcounter{table}{6}

\section{Bundle Adjustment Detail}

For patch $k$ with pixel coordinates $\mathbf{u}_k = [u_k, v_k]^T$ and depth $d_k$, we transform from image space to world coordinates. Define the normalized ray direction and coordinate transformations:
\begin{align}
\mathbf{n}_k &= \begin{bmatrix} (u_k - c_x)/f_x \\ (v_k - c_y)/f_y \\ 1 \end{bmatrix} \\
\mathbf{X}_k^c &= \pi^{-1}(\mathbf{u}_k, d_k) = d_k \mathbf{n}_k \\
\mathbf{X}_k^w &= \mathbf{R}_{t(k)} \mathbf{X}_k^c + \mathbf{t}_{t(k)}
\end{align}
where $\mathbf{X}_k^c$ and $\mathbf{X}_k^w$ are the 3D point in camera and world frames respectively, $t(k)$ denotes the frame index of patch $k$, and $\mathbf{T}_{t(k)} = [\mathbf{R}_{t(k)} \mid \mathbf{t}_{t(k)}]$ is the camera-to-world transformation. The scene coordinate residual is:
\begin{equation}
\mathbf{r}_k^{\text{xyz}} = w_k^{\text{xyz}} (\mathbf{X}_k^{\text{prior}} - \mathbf{X}_k^w) \in \mathbb{R}^3
\end{equation}
where $\mathbf{X}_k^{\text{prior}}$ is the scale-anchored coordinate prediction from the embedding $\mathbf{h}_k^{\text{xyz}}$, and $w_k^{\text{xyz}}$ is the confidence weight.

\subsection{Jacobian with Respect to Camera Pose}

For a Lie algebra perturbation $\boldsymbol{\tau}\in \mathfrak{se}(3)$ with translation $\boldsymbol{\rho}$ and rotation $\boldsymbol{\phi}$, the left Jacobian of the pose action is:

\begin{equation}
\mathbf{J}_{t(k)}^{\text{xyz}} = \frac{\partial \mathbf{r}_k^{\text{xyz}}}{\partial \boldsymbol{\tau}_{t(k)}} = 
-w_k^{\text{xyz}} 
\begin{bmatrix}
\mathbf{I}_{3\times3} & -[\mathbf{X}_k^w]_{\times}
\end{bmatrix} \in \mathbb{R}^{3 \times 6}
\end{equation}
where $[\mathbf{X}_k^w]_{\times}$ is the skew-symmetric matrix:
\begin{equation}
[\mathbf{X}_k^w]_{\times} = \begin{bmatrix}
0 & -Z & Y \\
Z & 0 & -X \\
-Y & X & 0
\end{bmatrix}
\end{equation}

\subsection{Jacobian with Respect to Depth}
From the backprojection definition $\mathbf{X}_k^c = d_k \mathbf{n}_k$, we have:
\begin{equation}
\frac{\partial \mathbf{X}_k^c}{\partial d_k} = \mathbf{n}_k
\end{equation}

Since $\mathbf{X}_k^w = \mathbf{R}_{t(k)} \mathbf{X}_k^c + \mathbf{t}_{t(k)}$ and the translation does not depend on depth:
\begin{equation}
\frac{\partial \mathbf{X}_k^w}{\partial d_k} = \mathbf{R}_{t(k)} \frac{\partial \mathbf{X}_k^c}{\partial d_k} = \mathbf{R}_{t(k)} \mathbf{n}_k
\end{equation}

Therefore:
\begin{equation}
\mathbf{J}_{d_k}^{\text{xyz}} = \frac{\partial \mathbf{r}_k^{\text{xyz}}}{\partial d_k} = -w_k^{\text{xyz}} \mathbf{R}_{t(k)} \mathbf{n}_k \in \mathbb{R}^{3 \times 1}
\end{equation}

\subsection{Hessian Matrix Construction}

In the Gauss-Newton framework, the Hessian matrix is approximated by the first-order term:
\begin{equation}
\mathbf{H}=\mathbf{J}^T \mathbf{W} \mathbf{J}
\end{equation}
where $\mathbf{W}$ is a diagonal weight matrix. We construct the weighted Jacobian matrices and residuals, and assemble them into a sparse block-structured linear system. Following the common practice in modern visual SLAM systems~\cite{teed2023dpvo,teed2021droid}, we solve this system using the Schur complement to efficiently marginalize the depth variables, yielding updates for the camera poses $\Delta \boldsymbol{\tau}$ and depth updates $\Delta \mathbf{d}$.

\section{Runtime Analysis}
\begin{figure*}[!t]
\centering
\includegraphics[width=1\linewidth]{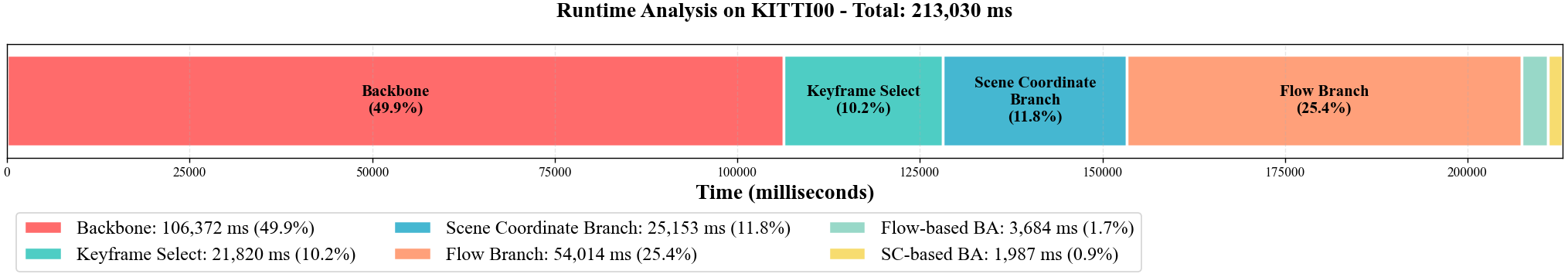}
\caption{\textbf{Component-wise Runtime analysis of the proposed method.}}
\label{fig:runtime_analysis}
\end{figure*}

We analyze the computational efficiency of our method by profiling the runtime of each component. As shown in Figure~\ref{fig:runtime_analysis}, the processing time is 213 seconds for the KITTI00~\cite{geiger2012kitti} sequence. The \textbf{backbone network} consumes the majority of the computation time (49.9\%), which is expected as it processes high-resolution images for feature extraction. The \textbf{flow branch} and \textbf{scene coordinate branch} account for 25.3\% and 11.8\% respectively.

Notably, our bundle adjustment with two complementary optimization objectives consumes only 2.7\% of the total time, demonstrating computational efficiency. Following the evaluation protocol of DPVO~\cite{teed2023dpvo}, we further benchmark our method on the EuRoC dataset~\cite{burri2016euroc} using 4090 GPU, achieving a processing speed of 36 FPS. The main bottleneck lies in the deep learning components, while our optimization adds minimal overhead, enabling real-time performance with high accuracy.

\section{Additional Results}

\noindent\textbf{Virtual KITTI.}
We evaluate robustness under diverse environmental conditions on the Virtual KITTI dataset~\cite{vkitti}, which includes various weather and lighting conditions such as clone, fog, morning, overcast, rain, and sunset (Table~\ref{tab:vkitti_complete}). Our method achieves the best overall performance across all scenes and conditions, significantly outperforming existing methods including DPVSLAM++~\cite{lipson2024dpvslam}.

\begin{figure*}[!t]
\centering
\includegraphics[width=1\linewidth]{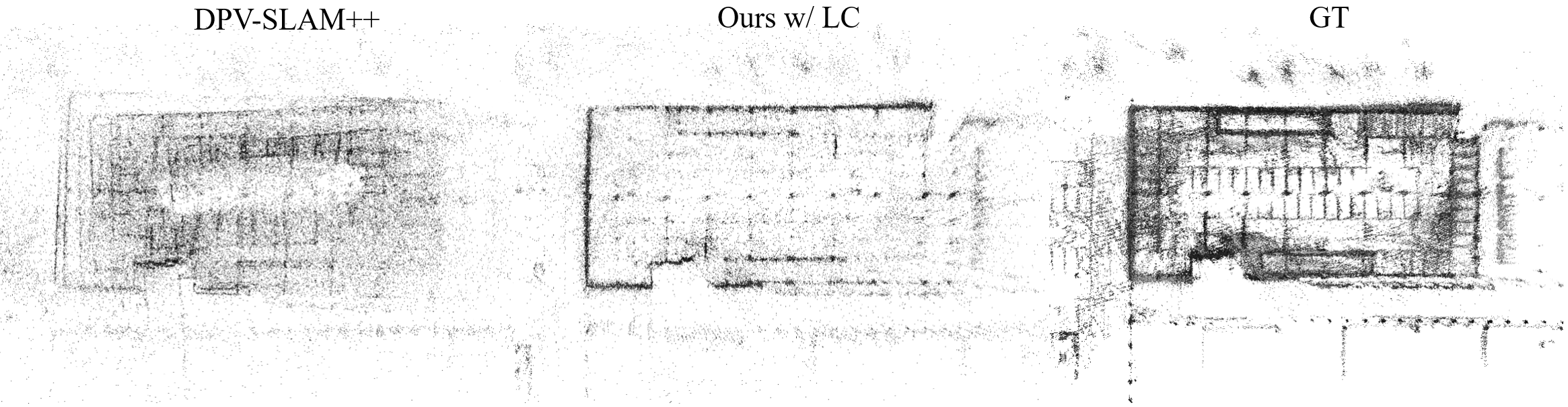}
\caption{\textbf{Comparison of reconstruction on the Parking Garage sequence from the 4 Seasons dataset.}}
\label{fig:fourseasonspcd}
\end{figure*}

\noindent\textbf{4 Seasons.}
On the Parking Garage sequence~\cite{fourseasons} featuring low-light and repetitive structures (Figure~\ref{fig:fourseasonspcd}), DPV-SLAM++ produces fragmented, layered point clouds due to pose drift, while our method generates coherent reconstruction closely matching the ground truth, demonstrating superior accuracy in challenging indoor scenarios.

\begin{table*}[t]
\resizebox{\textwidth}{!}{
\begin{tabular}{c|cccccc|cccccc}
\hline
\multicolumn{1}{l|}{}                     & \multicolumn{6}{c|}{\textbf{Scene 01}}                                                                                                                                                                                                           & \multicolumn{6}{c}{\textbf{Scene 02}}                                                                                                                                                                                                                    \\ \hline
{\color[HTML]{656565} \textit{Condition}} & {\color[HTML]{656565} \textit{Clone}} & {\color[HTML]{656565} \textit{Fog}} & {\color[HTML]{656565} \textit{Morning}} & {\color[HTML]{656565} \textit{Overcast}} & {\color[HTML]{656565} \textit{Rain}} & {\color[HTML]{656565} \textit{Sunset}} & {\color[HTML]{656565} \textit{Clone}}  & {\color[HTML]{656565} \textit{Fog}}    & {\color[HTML]{656565} \textit{Morning}} & {\color[HTML]{656565} \textit{Overcast}} & {\color[HTML]{656565} \textit{Rain}}   & {\color[HTML]{656565} \textit{Sunset}}   \\ \hline
DROID-SLAM~\cite{teed2021droid}                                & 1.027                                 & 1.868                               & 0.989                                   & 1.015                                    & 0.776                                & 1.145                                  & 0.098                                  & 0.040                                  & 0.049                                   & 0.047                                    & 0.036                                  & 0.112                                    \\
MASt3R-SLAM~\cite{murai2025mast3r}                               & TL                                    & TL                                  & TL                                      & TL                                       & TL                                   & TL                                     & TL                                     & TL                                     & TL                                      & TL                                       & TL                                     & TL                                       \\
CUT3R~\cite{wang2025cut3r}                                     & 43.304                                & 62.191                              & 50.608                                  & 38.735                                   & 51.548                               & 43.785                                 & 23.771                                 & 9.948                                  & 28.415                                  & 24.644                                   & 7.963                                  & 25.973                                   \\
VGGT-Long~\cite{deng2025vggt}                                 & 0.763                                 & 0.874                               & 0.928                                   & 0.670                                    & 1.799                                & 1.259                                  & 0.723                                  & 0.709                                  & 0.721                                   & 0.681                                    & 0.693                                  & 0.689                                    \\
DPVSLAM++~\cite{lipson2024dpvslam}                                 & {\ul 0.392}                           & \textbf{0.135}                      & {\ul 0.492}                             & \textbf{0.254}                           & {\ul 0.446}                          & \textbf{0.276}                         & \textbf{0.002}                         & {\ul 0.028}                            & \textbf{0.015}                          & {\ul 0.014}                              & {\ul 0.021}                            & {\ul 0.017}                              \\
Ours                                      & \textbf{0.269}                        & {\ul 0.137}                         & \textbf{0.487}                          & {\ul 0.265}                              & \textbf{0.346}                       & {\ul 0.285}                            & {\ul 0.014}                            & \textbf{0.025}                         & {\ul 0.022}                             & \textbf{0.012}                           & \textbf{0.017}                         & \textbf{0.016}                           \\ \hline
                                          & \multicolumn{6}{c|}{\textbf{Scene 06}}                                                                                                                                                                                                           & \multicolumn{6}{c}{\textbf{Scene 18}}                                                                                                                                                                                                                    \\ \hline
{\color[HTML]{656565} \textit{Condition}} & {\color[HTML]{656565} \textit{Clone}} & {\color[HTML]{656565} \textit{Fog}} & {\color[HTML]{656565} \textit{Morning}} & {\color[HTML]{656565} \textit{Overcast}} & {\color[HTML]{656565} \textit{Rain}} & {\color[HTML]{656565} \textit{Sunset}} & {\color[HTML]{656565} \textit{Clone}}  & {\color[HTML]{656565} \textit{Fog}}    & {\color[HTML]{656565} \textit{Morning}} & {\color[HTML]{656565} \textit{Overcast}} & {\color[HTML]{656565} \textit{Rain}}   & {\color[HTML]{656565} \textit{Sunset}}   \\ \hline
DROID-SLAM~\cite{teed2021droid}                                & 0.063                                 & 0.024                               & 0.030                                   & 0.051                                    & TL                                   & 0.020                                  & 2.478                                  & 2.032                                  & 1.893                                   & 2.332                                    & 2.550                                  & 1.943                                    \\
MASt3R-SLAM~\cite{murai2025mast3r}                               & TL                                    & TL                                  & TL                                      & TL                                       & TL                                   & TL                                     & TL                                     & TL                                     & TL                                      & TL                                       & TL                                     & TL                                       \\
CUT3R~\cite{wang2025cut3r}                                     & 0.836                                 & 0.408                               & 0.599                                   & 0.720                                    & 1.059                                & 1.013                                  & 19.440                                 & 8.628                                  & 6.720                                   & 20.212                                   & 16.777                                 & 31.119                                   \\
VGGT-Long~\cite{deng2025vggt}                                 & 0.365                                 & 0.543                               & 0.376                                   & 0.402                                    & 0.559                                & 0.382                                  & 1.651                                  & 0.797                                  & 1.288                                   & 1.256                                    & 1.648                                  & 1.740                                    \\
DPVSLAM++~\cite{lipson2024dpvslam}                                 & {\ul 0.055}                           & {\ul 0.054}                         & {\ul 0.050}                             & \textbf{0.069}                           & \textbf{0.055}                       & {\ul 0.077}                            & {\ul 0.449}                            & \textbf{0.016}                         & {\ul 0.179}                             & {\ul 0.217}                              & \textbf{0.151}                         & {\ul 0.207}                              \\
Ours                                      & \textbf{0.048}                        & \textbf{0.050}                      & \textbf{0.047}                          & {\ul 0.079}                              & {\ul 0.063}                          & \textbf{0.075}                         & \textbf{0.382}                         & {\ul 0.024}                            & \textbf{0.166}                          & \textbf{0.185}                           & {\ul 0.177}                            & \textbf{0.184}                           \\ \hline
                                          & \multicolumn{6}{c|}{\textbf{Scene 20}}                                                                                                                                                                                                           & \multicolumn{6}{c}{\textbf{AVG}}                                                                                                                                                                                                                         \\ \hline
{\color[HTML]{656565} \textit{Condition}} & {\color[HTML]{656565} \textit{Clone}} & {\color[HTML]{656565} \textit{Fog}} & {\color[HTML]{656565} \textit{Morning}} & {\color[HTML]{656565} \textit{Overcast}} & {\color[HTML]{656565} \textit{Rain}} & {\color[HTML]{656565} \textit{Sunset}} & {\color[HTML]{656565} \textit{01Avg.}} & {\color[HTML]{656565} \textit{02Avg.}} & {\color[HTML]{656565} \textit{06Avg.}}  & {\color[HTML]{656565} \textit{18Avg.}}   & {\color[HTML]{656565} \textit{20Avg.}} & {\color[HTML]{656565} \textit{All Avg.}} \\ \hline
DROID-SLAM~\cite{teed2021droid}                                & 3.592                                 & 5.079                               & 3.733                                   & 3.852                                    & 3.780                                & 4.907                                  & 1.137                                  & 0.064                                  & 0.038                                   & 2.205                                    & 4.157                                  & 1.520                                    \\
MASt3R-SLAM~\cite{murai2025mast3r}                               & TL                                    & TL                                  & TL                                      & TL                                       & TL                                   & TL                                     & TL                                     & TL                                     & TL                                      & TL                                       & TL                                     & TL                                       \\
CUT3R~\cite{wang2025cut3r}                                     & 129.498                               & 76.962                              & 117.948                                 & 114.512                                  & 66.700                               & 116.529                                & 48.362                                 & 20.119                                 & 0.772                                   & 17.149                                   & 103.692                                & 38.019                                   \\
VGGT-Long~\cite{deng2025vggt}                                 & 9.655                                 & 8.185                               & 6.345                                   & 4.564                                    & 6.499                                & 4.85                                   & 1.049                                  & 0.703                                  & 0.438                                   & 1.397                                    & 6.683                                  & 2.054                                    \\
DPVSLAM++~\cite{lipson2024dpvslam}                                 & {\ul 0.924}                           & {\ul 2.257}                         & {\ul 0.648}                             & {\ul 0.648}                              & {\ul 1.242}                          & {\ul 0.9176}                           & {\ul 0.332}                            & \textbf{0.016}                         & \textbf{0.060}                          & {\ul 0.203}                              & {\ul 1.106}                            & {\ul 0.344}                              \\
Ours                                      & \textbf{0.743}                        & \textbf{1.544}                      & \textbf{0.479}                          & \textbf{0.593}                           & \textbf{0.876}                       & \textbf{0.794}                         & \textbf{0.298}                         & {\ul 0.018}                            & \textbf{0.060}                          & \textbf{0.186}                           & \textbf{0.838}                         & \textbf{0.280}                           \\ \hline
\end{tabular}
}
\caption{\textbf{Comparison on the Virtual KITTI Dataset of ATE RMSE $\downarrow$ (m).}  The \textbf{best} and {\ul second-best} results are marked.}
\label{tab:vkitti_complete}
\end{table*}

\clearpage
{
    \small
    \bibliographystyle{ieeenat_fullname}
    \bibliography{main}
}


\end{document}